%% file: main.tex
\documentclass[11pt,a4paper]{article}
\usepackage[hyperref]{acl2020}
\usepackage{times}
\usepackage{latexsym}

\usepackage{microtype}

\aclfinalcopy %
\def\aclpaperid{3194} %

\usepackage{amsmath}
\usepackage{amssymb}
\usepackage{booktabs}
\usepackage{subcaption,graphicx}
\usepackage{todonotes}
\usepackage{color,soul}

\input{mathdefs.tex}

\DeclareMathOperator*{\argmin}{arg\;min}

\newif\ifdraft
\drafttrue

\newif\ifedit
\edittrue

\title{Discrete Latent Variable Representations \\ for Low-Resource Text Classification}

\author{
Shuning Jin$^1$\thanks{\hspace{0.5em}Work done as an intern at Toyota Technological Institute at Chicago.}\qquad
Sam Wiseman$^2$\qquad
Karl Stratos$^1$\qquad
Karen Livescu$^2$\\
$^1$Rutgers University\quad 
$^2$Toyota Technological Institute at Chicago\\
\small{\texttt{shuning.jin@rutgers.edu},\;
\texttt{\{swiseman,klivescu\}@ttic.edu},\;
\texttt{stratos@cs.rutgers.edu} }
}

\date{}

\begin{document}
\maketitle
\begin{abstract}
While much work on deep latent variable models of text uses continuous latent variables, discrete latent variables are interesting because they are more interpretable and typically more space efficient. We consider several approaches to learning discrete latent variable models for text in the case where exact marginalization over these variables is intractable. We compare the performance of the learned representations as features for low-resource document and sentence classification.
Our best models outperform the previous best reported results with continuous representations in these low-resource settings, while learning significantly more compressed representations. Interestingly, we
find that an amortized variant of Hard EM performs particularly well in the lowest-resource regimes.%
\footnote{Code available on GitHub:\,\,\,\url{https://github.com/shuningjin/discrete-text-rep}}
\end{abstract}

\section{Introduction}
Deep generative models with latent variables have become a major focus of NLP research over the past several years.
These models have been used both for generating text~\citep{bowman-etal-2016-generating} and as a way of learning latent representations of text for downstream tasks~\citep{yang2017improved,gururangan2019variational}. Most of this work has modeled the latent variables as being continuous, that is, as vectors in $\reals^d$, 
in part due to the simplicity of performing inference over (certain) continuous latents using variational autoencoders and the reparameterization trick~\citep{Kingma2014,Rezende2014}.

At the same time, deep generative models with \textit{discrete} latent variables are attractive because the latents are arguably more interpretable, and because they lead to significantly more compressed representations: A representation consisting of $M$ floating point values conventionally requires $M \times 32$ bits, whereas $M$ integers in $\{1, \ldots, K\}$ requires only $M \times \log_2 K$ bits.

Unfortunately, discrete latent variable models have a reputation for being more difficult to learn.
We conduct a thorough comparison of several popular methods for learning such models, all within the framework of maximizing the evidence lower bound (ELBO)
on the training data.
In particular, we compare learning such models either with a Vector Quantized-VAE %
\citep[VQ-VAE]{oord2017neural},
a more conventional VAE with discrete latent variables~\citep{jang2017categorical,maddison2017concrete}, or with an amortized version of ``Hard''  or ``Viterbi'' Expectation Maximization~\citep{brown1993mathematics}, which to our knowledge has not been explored to date. We consider both models where the latents are local (i.e., per token) and
where they are global (i.e., per sentence);
we assess the quality of these learned discrete representations as features for a low-resource text classifier, as suggested by \citet{gururangan2019variational}, and in a nearest neighbor-based retrieval task. 

Our classification experiments distinguish between (1) the setting where the classifier must consume only the discrete representation associated with each sentence (i.e., the discrete assignment that maximizes the approximate posterior), and (2) the setting where the classifier may consume the \textit{embeddings} of this discrete representation learned by the VAE encoder. Note that the former setting is more flexible, since we need only store a sentence's discrete representation, and are therefore free to use task-specific (and possibly much smaller) architectures for classification. In case~(1),
we are able to effectively match the performance of \citet{gururangan2019variational} and other baselines;
in case~(2), we outperform them.
Our experiments also suggest that Hard EM performs particularly well in case~(1) when there is little supervised data, and that VQ-VAE struggles in this setting.

\section{Related Work}

Our work builds on recent advances in discrete representation learning and its applications. 
In particular, we are inspired by recent success with VQ-VAEs outside NLP
\citep{oord2017neural, razavi2019generating}. 
These works show that we can generate realistic speech and image samples from discrete encodings, which better align with symbolic representations that humans seem to work with (e.g., we naturally encode continuous speech signals into discrete words).
Despite its success in speech and vision, VQ-VAE has not been considered as much in NLP. One exception is the translation model of \citet{kaiser2018fast} that encodes a source sequence into discrete codes using vector quantization. But their work focuses on making inference faster,
by decoding the target sequence from the discrete codes non-autoregressively.
To our knowledge, we are the first that explores general text representations induced by VQ-VAEs for semi-supervised and transfer learning in NLP.

In addition to exploring the viability of VQ-VAEs for text representation learning, an important part of this paper is a systematic comparison between different discretization techniques.
Gumbel-Softmax \citep{jang2017categorical,maddison2017concrete} is a popular choice that has been considered for supervised text classification \citep{chen2019smaller} and dialog generation \citep{zhao2018unsupervised}.
In the binary latent variable setting, straight-through estimators are often used \citep{dong2019document}.
Another choice is ``continuous decoding'' which takes a convex combination of latent values to make the loss differentiable \citep{al-shedivat-parikh-2019-consistency}.
Yet a less considered choice is Hard EM~\citep{brown1993mathematics,de1995lexical,spitkovsky2010viterbi}.
A main contribution of this work is a thorough empirical comparison between such different choices in a controlled setting. 

To demonstrate the usefulness of our models, we focus on improving low-resource classification performance by pretraining on unlabeled text.
Previous best results are obtained with continuous latent-variable VAEs,
e.g., VAMPIRE \citep{gururangan2019variational}.
We show that our discrete representations outperform these previous results while being significantly more lightweight.

\section{Background}
We consider generative models of a sequence $x = x_{1:T}$ of $T$ word tokens. We
assume our latents to be a sequence $z = z_{1:L}$ of $L$ discrete latent vectors, each taking a value in $\{1, \ldots, K\}^M$; that is, $z \in \{1, \ldots, K\}^{M \times L}$. As is common in VAE-style models of text, we model the text autoregressively, and allow arbitrary interdependence between the text and the latents. That is, we have $p(x, z \param \btheta) = p(z) \times \prod_{t=1}^T p(x_t \given x_{<t}, z \param \btheta)$, where $\btheta$ are the generative model's parameters. We
further assume $p(z)$ to be a fully factorized, uniform prior: $p(z) = \frac{1}{K^{ML}}$. 

Maximizing the marginal likelihood of such a model will be intractable for moderate values of $K$, $M$, and $L$.
So we consider learning approaches that maximize the ELBO~\citep{jordan1999introduction} in an amortized way~\citep{Kingma2014,Rezende2014}:
\begin{align*}
\ELBO(\btheta, \bphi) = \E_{q(z \given x \param \bphi)} \left[ \log \frac{p(x, z \param \btheta)}{q(z \given x \param \bphi)} \right],
\end{align*}
where $q(z \given x \param \bphi)$ is the approximate posterior given by an inference or encoder network with parameters $\bphi$. The approaches we consider differ in terms of how this approximate posterior $q$ is defined.

\paragraph{Mean-Field Categorical VAE (CatVAE)}
A standard Categorical VAE
parameterizes the approximate posterior as factorizing over categorical distributions that are independent given $x$. We therefore maximize:

\vspace*{-0.5cm}
{%
\begin{align*}
& \E_{q(z \given x \param \bphi)} \left[ \log p(x \given z \param \btheta) \right] - \sum_{m, l} \KL(q_{ml} || p_{ml}) \\
&= \E_{q(z \given x \param \bphi))} \left[ \log p(x \given z \param \btheta) \right]\\
&\quad + \sum_{m,l} H(q_{ml}) - ML \log K,
\end{align*}
}

\vspace*{-0.3cm}
\noindent where $q(z \given x \param \bphi) {=} \prod_{m=1}^M \prod_{l=1}^L q_{ml}(z_{ml} \given x \param \bphi)$, $p_{ml} = 1/K$, and $H$ is the entropy.

We approximate the expectation above by sampling from the $q_{ml}$, and we use the straight-through gradient estimator~\citep{bengio2013estimating, jang2017categorical} to compute gradients with respect to $\bphi$. We find this approach to be more stable than using the REINFORCE~\citep{Williams1992} gradient estimator, or a
Concrete~\citep{maddison2017concrete,jang2017categorical} approximation to categorical distributions.
Specifically, we sample from a categorical distribution using the Gumbel-Max trick~\citep{maddison2014sampling} in the forward pass, and approximate the gradient using softmax with a small temperature. This approach is also referred to as straight-through Gumbel-Softmax
\citep{jang2017categorical}.

\paragraph{VQ-VAE}
A VQ-VAE \citep{oord2017neural, razavi2019generating} can also be seen as maximizing the ELBO, except the approximate posterior is assumed to be a point mass given by

\vspace*{-0.3cm}
{%
\begin{align*} q_{ml}(z_{ml}|x) = 
\begin{cases} 
      1 & \text{if } z_{ml} = \hat{z}_{ml} \\
      0 & \text{otherwise } %
   \end{cases},
\end{align*}
}

\noindent where 
{%
\begin{align} \label{eq:quantize}
    \hat{z}_{ml} = \argmin_{j \in \{1, \ldots, K\}} ||\bolde^{(m)}_j - \enc(x)_{ml}||_2,
\end{align}
}

\noindent and $\bolde^{(m)}_j \in \reals^d$ is an embedding of the $j^\textrm{th}$ discrete value $z_{ml}$ can take on, and $\enc(x)_{ml} \in \reals^d$ is an encoding corresponding to the $ml^\textrm{th}$ latent given by an encoder network. These $\bolde^{(m)}_j$ embedding vectors are often referred to as a VQ-VAE's ``code book''.
In our setting, a code book is shared across latent vectors. 

VQ-VAEs are typically learned by maximizing the ELBO assuming degenerate approximate posteriors as above, plus two terms that encourage the encoder embeddings and the ``code book'' embeddings to become close. In particular, we attempt to maximize the objective:

\vspace*{-0.3cm}
{ %
\begin{align} \label{eq:vqloss}
    \log p(x \given \hat{z}) &- \sum_{m,l} ||\mathrm{sg}(\enc(x)_{ml}) - \bolde^{(m)}_{\hat{z}_{m,l}}||_{2}^{2} \\
    &-\beta\sum_{m,l}  ||\enc(x)_{ml} - \mathrm{sg}(\bolde^{(m)}_{\hat{z}_{m,l}})||_{2}^{2}, \nonumber
\end{align}
}

\noindent where $\mathrm{sg}$ is the stop-gradient operator, and $\hat{z} = \hat{z}_{1:L}$ is the sequence of minimizing assignments $\hat{z}_{m,l}$ for each $\enc(x)_{ml}$. The loss term following the $\beta$ is known as the ``commitment loss''.
Gradients of the likelihood term with respect to $\enc(x)$ are again estimated with the straight-through gradient estimator.

\paragraph{Hard EM}
We train with an amortized form of Hard EM. First we define a relaxed version of $z$, $\tilde{z}$, where each 
$\tilde{z}_{ml}$ is a $\softmax$ over $K$ outputs (rather than a hard assignment) and is produced by an inference network with parameters $\bphi$.%
\footnote{Note this assumes our generative model can condition on such a relaxed latent variable.}
In the E-Step, we take a small, constant number of gradient steps to maximize $\log p(x \given \tilde{z} \param \btheta)$ with respect to $\bphi$ (for a fixed $\btheta$). In the M-Step, we take a single gradient step to maximize $\log p(x \given \hat{z} \param \btheta)$ with respect to $\btheta$, where $\hat{z}$ contains the element-wise argmaxes of $\tilde{z}$ as produced by the inference network (with its most recent parameters $\bphi$). Thus, Hard EM can also be interpreted as maximizing the (relaxed) ELBO. We also note that taking multiple steps in the hard E-step somewhat resembles the recently proposed aggressive training of VAEs~\citep{he2019lagging}.

\section{Models and Architectures}\label{sec:models}

\begin{figure}[t!] 
\hspace*{-0.2cm}
\includegraphics[scale=0.40]{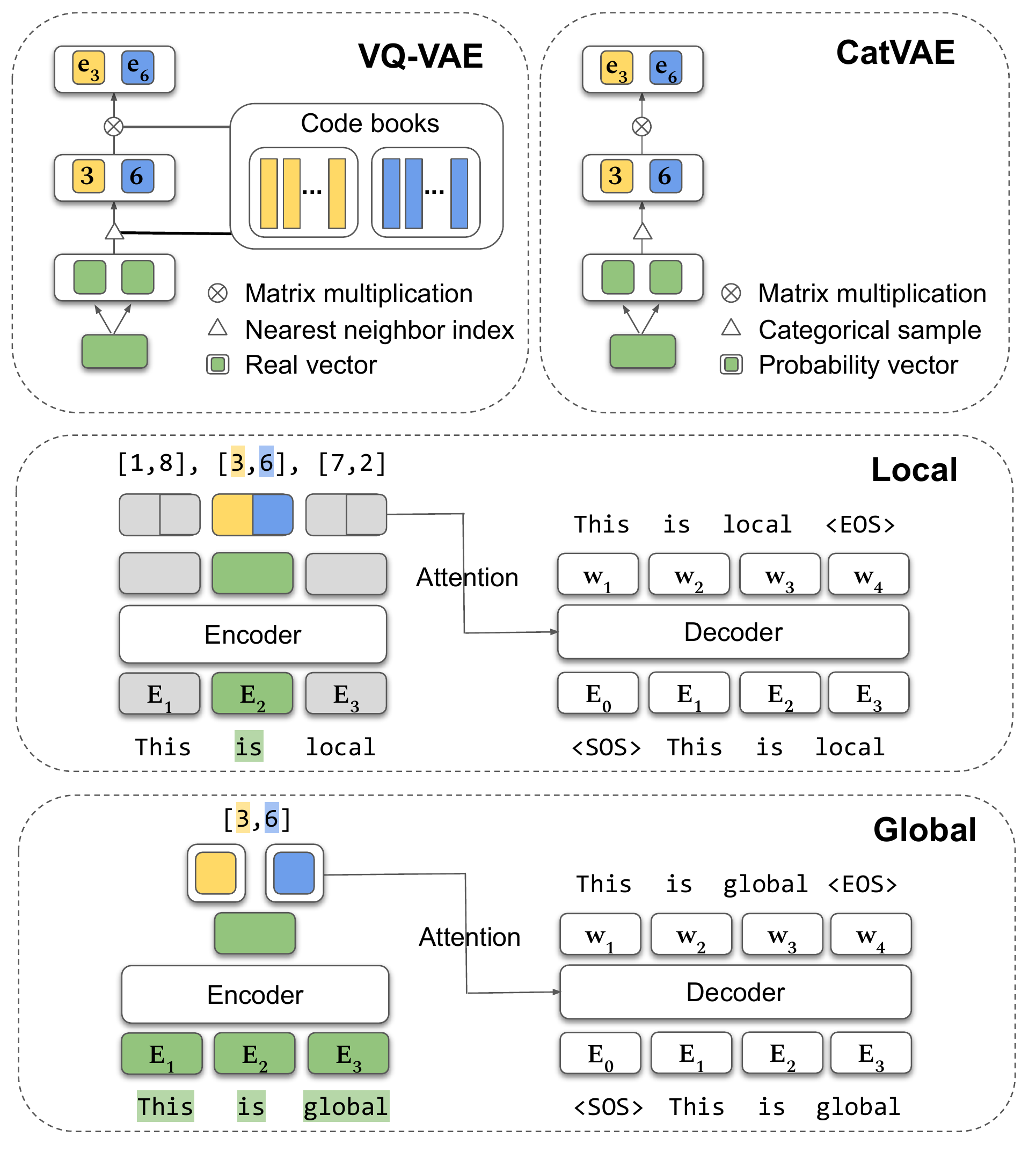}
\caption{Discrete VAE architectures with $M=2$. The  \textbf{Local} (middle) and \textbf{Global} (bottom) models are two different encoder-decoder setups. The top row shows the procedure of converting continuous output from encoder into discrete input to decoder by drawing discrete samples: \textbf{VQ-VAE} (top left) draws samples from point mass distributions using nearest neighbor lookup from the code books; \textbf{CatVAE} (top right) samples from categorical distributions directly.
}
\label{fig:architecture}
\end{figure}

Recall that the latent sequence is $z = z_{1:L}$, where $z_l \in \{1, \ldots, K\}^M$.
We consider two generative models $p(x \given z \param \btheta)$, one where $L=T$ and one where $L = 1$.
Each latent in the former model corresponds to a word, and so we refer to this as a ``local'' model,
whereas in the second model we view the latents as being ``global'', since there is one latent vector for the whole sentence. We use the following architectures for our encoders and decoder, as illustrated in Figure~\ref{fig:architecture}.

\subsection{Encoder}

The encoder (parameterized by $\bphi$) maps an example $x$ to the parameters of an approximate posterior distribution. Our encoder uses a single-layer Transformer~\citep{vaswani2017attention} network to map $x = x_{1:T}$ to a sequence of $T$ vectors $\boldh_1, \ldots, \boldh_T$, each in $\reals^{d}$.

\paragraph{Mean-Field Categorical VAE}
For the local model, we obtain the parameters of each categorical approximate posterior $q_{mt}$ as $\softmax(\boldW_m \, \boldh_t)$, where each $\boldW_m \in \reals^{K \times d}$ is a learned projection. For the global model, we obtain the parameters of each categorical approximate posterior $q_{m1}$ as
$\softmax\left(\frac{\sum_t \boldW_m \; \boldh_t}{T}\right)$; that is, we pass token-level $\boldh_t$ vectors through learned projections $\boldW_m$, followed by mean-pooling.

\paragraph{VQ-VAE} 

For the local model, let
$\tilde{d} = {d}/{M}$. 
We obtain $\enc(x)_{mt}$, the encoding of the $mt^{\textrm{th}}$ latent variable, as $\boldh_{t,(m-1)\tilde{d}:m\tilde{d}}$, following \citet{kaiser2018fast}. That is, we take the $m^{\textrm{th}}$ $\tilde{d}$-length subvector of $\boldh_t$.
For the global model, let $\tilde{d} = d$. We first project $\boldh_t$ to $\reals^{Md}$, mean-pool,
and obtain $\enc(x)_{m1}$ by taking the $m^{\textrm{th}}$ $\tilde{d}$-length subvector of the resulting pooled vector. A VQ-VAE also requires learning a code book, and we define $M$ code books $\mathbf{E}^{(m)} = [{\bolde_1^{(m)}}^\top; \ldots ; {\bolde_K^{(m)}}^\top] \in$ $\reals^{K \times \tilde{d}}$ .

\paragraph{Hard EM}
We use the same encoder architecture as in the mean-field Categorical VAE case. Note, however, that we do not sample from the resulting categorical distributions. Rather, the $\softmax$ distributions are passed directly into the decoder.

\subsection{Decoder}
In the case of the mean-field Categorical VAE, we obtain a length-$L$ sequence of vectors $z_l \in \{1, \ldots, K\}^{M}$ after sampling from the approximate posteriors. For the VQ-VAE, on the other hand, we obtain the sequence of $\hat{z}_l$ vectors by taking the indices of the closest code book embeddings, as in Equation~\eqref{eq:quantize}. 

In both cases, the resulting sequence of discrete vectors is embedded and consumed by the decoder. In particular, when learning with a VQ-VAE, the embedding of $\hat{z}_{ml}$ is simply $\bolde^{(m)}_{\hat{z}_{ml}}$, whereas for the Categorical VAE each discrete latent is embedded using a trained embedding layer.
In the local model, when $M>1$, we concatenate the $M$ embeddings to form a single real vector embedding for the $l^{ \textrm{th} }$ latent variable.
In the global model, we use the $M$ embeddings directly.
This resulting sequence of 
$T$ or $M$ real vectors
is then viewed as the source side input for a standard 1-layer Transformer encoder-decoder model~\citep{vaswani2017attention}, which decodes $x$ using causal masking.

As above, for Hard EM, we do not obtain a sequence of discrete vectors from the encoder, but rather a sequence of $\softmax$ distributions. These are multiplied into an embedding layer, as in the Categorical VAE case, and fed into the Transformer encoder-decoder model.

\section{Evaluating Latent Representations}\label{sec:evaluating}

Similar to~\citet{gururangan2019variational}, we evaluate the learned latent representations by using them as features in a text classification system. We are in particular interested in using latent representations learned on unlabeled text to help improve the performance of classifiers trained on a small amount of labeled text. %
Concretely, we compare different discrete latent variable models in following steps:

\begin{enumerate}
    \item 
    Pretraining an encoder-decoder model on in-domain unlabeled text with an ELBO objective, with early stopping based on validation perplexity.
    \item Fixing the encoder to get discrete latents for the downstream classification task, and training a small number of task-specific parameters on top, using varying amounts of labeled data. As noted in the introduction, we consider both reembedding these latents from scratch, or using the embeddings learned by the encoder.
\end{enumerate}

\subsection{Tasks and Datasets}
The datasets we use for classification are AG News, DBPedia, and Yelp Review Full~\citep{zhang2015characterlevel}, which correspond to predicting news labels, Wikipedia ontology labels, and the number of Yelp stars, respectively. The data details are summarized in Table~\ref{tab:datasets}.  For all datasets, we randomly sample 5,000 examples as development data. To evaluate the efficiency of the latent representation in low-resource settings, we train the classifier with varying numbers of labeled instances: 200, 500, 2500, and the full training set size (varies by dataset). We use accuracy as the evaluation metric. 

In preprocessing, we space tokenize, lowercase, and clean the text as in \citet{kim2014convolutional}, and then truncate each sentence to a maximum sequence length of 400. For each dataset, we use a vocabulary of the 30,000 most common words. 

\subsection{Transfer Paradigm}

When transferring to a downstream classification task, we freeze the pretrained encoder and add a lightweight classifier on top, viewing each sentence as an $L$-length sequence of vectors in $\{1, \ldots, K\}^M$, as described in Section~\ref{sec:models}.
For instance, the sentence (from the DBPedia dataset) \textit{``backlash is a 1986 australian film directed by bill bennett''} is encoded as
\textsc{[90, 114, 30, 111]} under a global model with $M=4$, and as
\textsc{[[251, 38], [44, 123], [94, 58], [228, 53], [88, 55], [243, 43], [66, 236], [94, 72], [172, 61], [236, 150]]} under a local model with $M=2$.

As noted in the introduction, we consider two ways of embedding the integers for consumption by a classifier. We either (1) learn a new task-specific embedding space $\mathbf{E}^{(m)}_\text{task}$ (i.e., reembedding) or (2) use the fixed embedding space $\mathbf{E}^{(m)}$ from pretraining. The first setting
allows us to effectively replace sentences with their lower dimensional discrete representations,
and learn a classifier on the discrete representations from scratch. In the local model, we obtain token-level embedding vectors by concatenating the $M$ subvectors corresponding to each word.
The resulting embeddings are
either averaged, or fed to a Transformer and then averaged, and
finally fed into a linear layer followed by a $\softmax$.

\begin{table}[t]
\centering \fontsize{9.6}{12}\selectfont \setlength{\tabcolsep}{0.5em}
\input{tables/tasks.tex}
\caption{The number of classes and the numbers of examples in each data subset, for the classification tasks.} 
\label{tab:datasets}
\end{table}

\section{Experimental Details} 
\subsection{Baselines}
We first experiment with three common text models: CBOW~\citep{mikolov2013distributed}, bidirectional LSTM~\citep{Hochreiter1997}, and a single-layer Transformer encoder. We find CBOW (with 64-dimensional embeddings) to be the most robust in settings with small numbers of labeled instances,
and thus report results only with this baseline
among the three. Further, we compare to VAMPIRE \citep{gururangan2019variational},
a framework of pretraining VAEs for text classification using continuous latent variables. We pretrain VAMPIRE models on in-domain text for each dataset with 60 random hyperparameter search (with same ranges as specified in their Appendix A.1), and select best models based on validation accuracy in each setting.

\subsection{Hyperparameters}
In our experiments, we use Transformer layers with $d_\text{model}=64$. For optimization, we use Adam \citep{kingma2014adam}, either with a learning rate of 0.001 or with the inverse square-root schedule defined in \citet{vaswani2017attention} in pretraining. We use a learning rate of 0.0003 in classification. We tune other hyperparameters with random search and select the best settings based on validation accuracy.
For the latent space size, we choose $M$ in $\{1,2,4,8,16\}$ and $K$ in $\{128, 256, 512, 1024, 4096\}$. Model specific hyperparameters are introduced below.

\subsection{VQ-VAE}
In VQ-VAE, an alternative to the objective in Equation~\eqref{eq:vqloss} is to remove its second term, while using an auxiliary dictionary learning algorithm with exponential moving averages (EMA) to update the embedding vectors \citep{oord2017neural}.  We tune whether to use EMA updates or not. Also, we find small $\beta$ for commitment loss to be beneficial, and search over $\{0.001, 0.01, 0.1\}$.

\begin{figure*}[t!]
\centering
\includegraphics[scale=0.73]{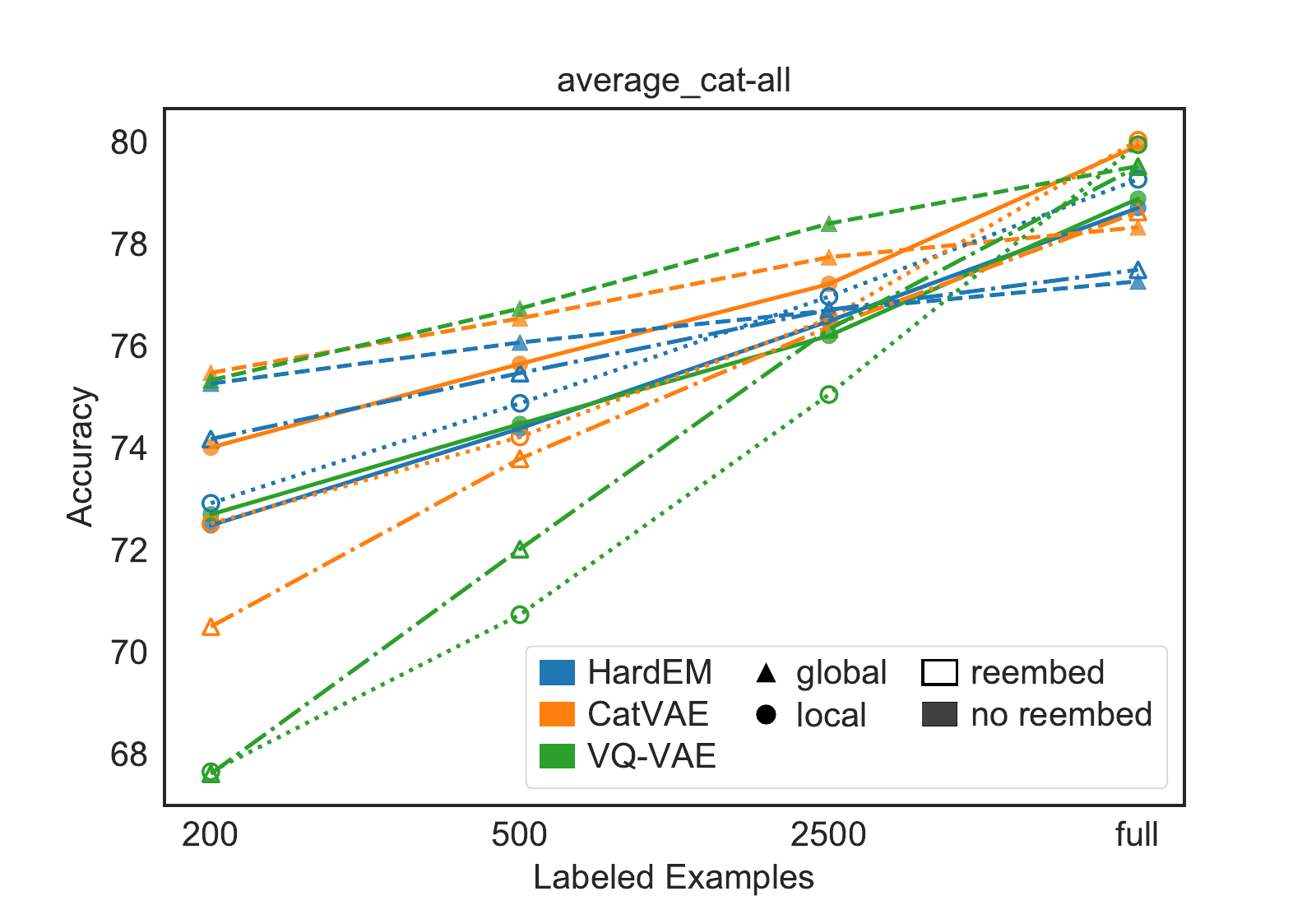}
\caption{The accuracies obtained by Hard EM, Categorical VAE, and VQ-VAE representations, averaged over the AG News, DBPedia, and Yelp Full development datasets, for different numbers of labeled training examples.
Triangular and circular markers correspond to global and local models, respectively. Unshaded and shaded markers correspond to reembedding from scratch and using encoder embeddings, respectively.
}
\label{fig:devresults}
\end{figure*}

\subsection{Mean-Field Categorical VAE}

We find that using the discrete analytic KL divergence term directly in the ELBO objective leads to posterior collapse. The KL term vanishes to 0 and the $q_{ml}$ distributions converge to the uniform priors. To circumvent this, we modify the KL term to be $\mathrm{max}( \text{KL}, \lambda)$. This is known as Free Bits \citep{kingma2016improving, li-etal-2019-surprisingly}, which ensures that the latent variables encode a certain amount of information by not penalizing the KL divergence when it is less than $\lambda$. We set $\lambda = \gamma ML \log K$, where $\gamma$ is a hyperparameter between 0 and 1. That is, we allocate a ``KL budget'' as a fraction of $ML \log K$, which is the upper bound of KL divergence between $ML$ independent categorical distributions and uniform prior distributions. Since in this case $\KL(q_{ml}(z_{ml} \given x)||p_{ml}(z_{ml})) = \log K - H[q_{ml}(z_{ml}\given x)]$, this is equivalent to thresholding $H[q_{ml}(z_{ml} \given x)]$ by $(1-\gamma) \log K$. We experiment with $\gamma \in \{0.2, 0.4, 0.6, 0.8, 1\}$.%
\footnote{Note that when $\gamma \ge 1$ the VAE reduces to an autoencoder.}

\subsection{Hard EM}
We vary the number of gradient steps in the E-step in $\{1,3\}$. At evaluation time, we always take the $\mathrm{argmax}$ of $\tilde{z}$ to get a hard assignment.

\begin{figure}[t!]
\includegraphics[scale=0.55]{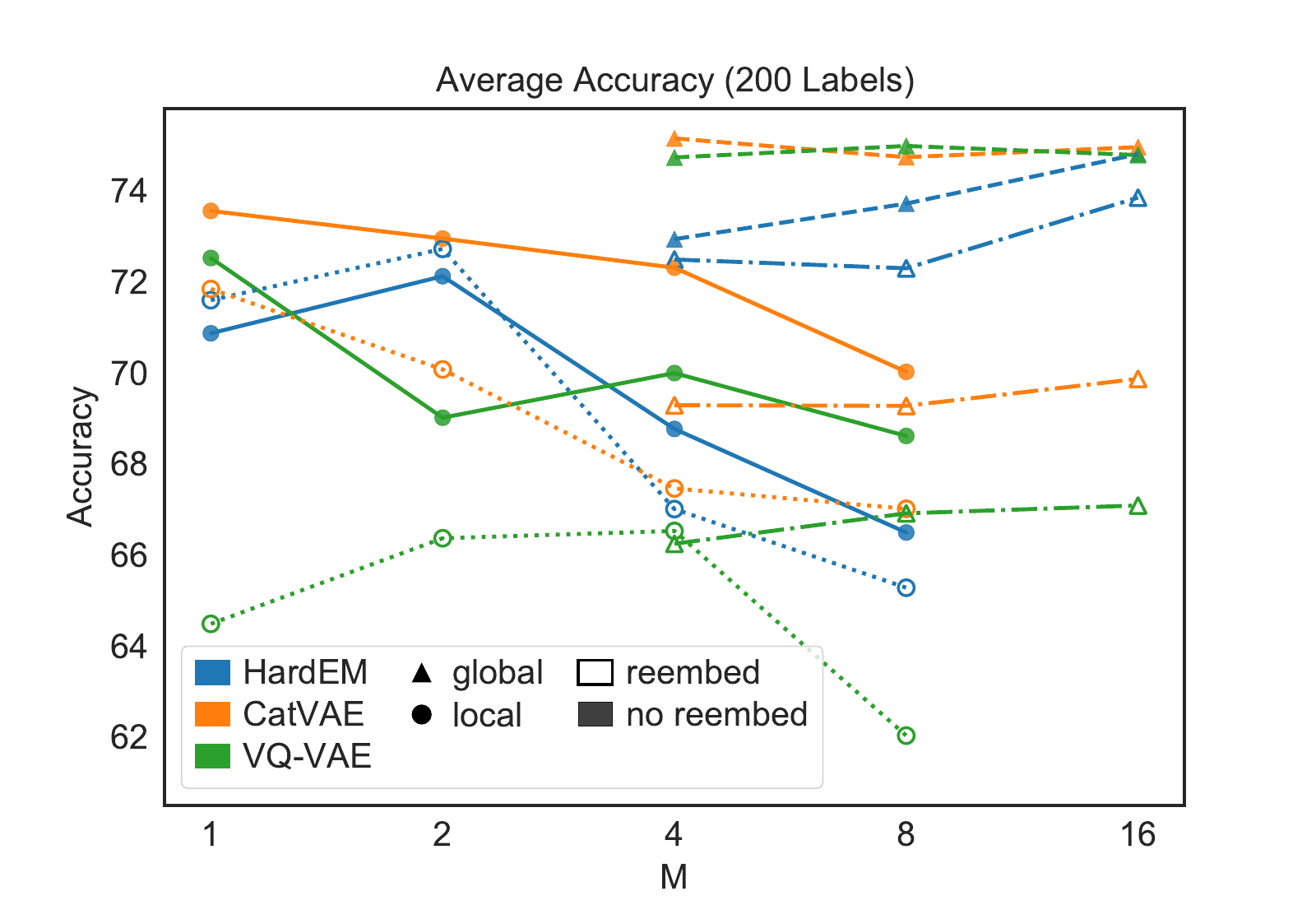}
\caption{The averaged accuracies obtained from using Hard EM, Categorical VAE, and VQ-VAE representations and 200 labeled examples, for different $M$ values.
}
\label{fig:mfig}
\end{figure}

\section{Results}

In Figure~\ref{fig:devresults}, we compare the accuracy obtained by the representations from our Hard EM, Categorical VAE, and VQ-VAE models,
averaged over the development datasets of AG News, DBPedia, and Yelp Full.
In particular, we plot the best accuracy obtained over all hyperparameters (including $M$) for different numbers of labeled examples;
we distinguish between local and global models, and between when the discrete representations are reembedded from scratch and when the encoder embeddings are used.

We see that using the encoder embeddings typically outperforms reembedding from scratch, and that global representations tend to outperform local ones, except in the full data regime. Furthermore, we see that the Categorical VAE and VQ-VAE are largely comparable on average, though we undertake a finer-grained comparison by dataset in
Appendix~\ref{sec:supplemental}. Perhaps most interestingly, we note that when reembedding from scratch, Hard EM significantly outperforms the other approaches in the lowest data regimes (i.e., for 200 and 500 examples). In fact, Hard EM allows us to match the performance of the best previously reported results even when reembedding from scratch; see Table~\ref{tab:test}.

Table~\ref{tab:hypers} shows
the best combinations of model and hyperparameters when training with 200 labeled examples on AG News. These settings were
used in obtaining the numbers in Figure~\ref{fig:devresults}, and are largely stable across datasets.

\begin{table}[t!]
\centering
\input{tables/hypers.tex}
\caption{Best methods and settings of $K$ and $M$ when training on 200 labeled examples of the AG News corpus and evaluating on the development set. The ``(re)'' affix indicates that latent variables are reembedded from scratch.
}
\label{tab:hypers}
\end{table}

\begin{table}[t!]
\centering
\small
\input{tables/test.tex}
\caption{Test accuracy results by dataset and by the number of labeled examples used in training. The scores are averages over five random subsamples, with standard deviations in parentheses and column bests in \textbf{bold}.
VAMPIRE$^\star$ for AG News is reported by \citet{gururangan2019variational} and VAMPIREs are from our experiments.
}
\label{tab:test}
\end{table}

In Figure~\ref{fig:mfig}, we compare the average accuracy of our local and global model variants trained on 200 labeled examples, as we vary $M$. 
When reembedding, local representations tend to improve as we move from $M=1$ to $M=2$, but not significantly after that. When reembedding global representations, performance increases as $M$ does. Unsurprisingly, when not reembedding, $M$ matters less.

Finally, we show the final accuracies obtained by our best models on the test data of each dataset in Table~\ref{tab:test}.
We see that on all datasets when there are only 200 or 500 labeled examples, our best model outperforms VAMPIRE and the CBOW baseline, and our models that reembed the latents from scratch match or outperform VAMPIRE. As noted in Table~\ref{tab:hypers}, it is Hard EM that is particularly performant in these settings.

\begin{table*}[t]
\centering %
\small
\setlength{\tabcolsep}{0.5em}
\input{tables/examples.tex}

\caption{Examples of sentence-level ($M=4$, $K=256$) clusters on AG News.}
\label{tab:ex}
\end{table*}

\begin{table*}[t]
\centering %
\small
\setlength{\tabcolsep}{0.5em}
\input{tables/examples2.tex}
\caption{Word-level ($M=1$, $K=1024$) clusters on AG News. We take the majority cluster for each word for illustration purposes.}
\label{tab:ex2}
\end{table*}

\section{Analysis and Discussion}

\subsection{Qualitative analysis}
To gain a better understanding of what the learned clusters represent, we examine their patterns on the AG News dataset labeled with four classes. Since VQ-VAEs and Categorical VAEs
exhibit similar patterns, we focus on the latter model.  

Tables~\ref{tab:ex} and~\ref{tab:ex2} show examples of sentence- and word-level clusters, respectively, induced by Categorical VAEs. 
The sentence-level model encodes each document into $M=4$ latents, each taking one of $K=256$ integers. The word-level model encodes each word into $M=1$ latent taking one of $K=1024$ integers. Since a word can be assigned multiple clusters, we take the majority cluster for illustration purposes. 

We see that clusters correspond to topical aspects of the input (either a document or a word). In particular, in the sentence-level case, documents in the same cluster often have the same ground-truth label. We also find that each of $M$ latents independently corresponds to topical aspects (e.g., $z_1 = 65$ implies that the topic has to do with technology); thus, taking the combination of these latents seems to make the cluster ``purer''. 
The word-level clusters are also organized by topical aspects (e.g., many words in cluster 510 are about modern conflicts in the Middle East). 

\subsection{Effect of Alternating Optimization}

\begin{table}[t!]
\centering
\small
\input{tables/alternating.tex}
\caption{Effect of alternating optimization on AG News classification with 200 and 500 labels. The ``(re)'' affix denotes reembedding. Accuracies are on development set with column highs in \textbf{bold}.}
\label{tab:alternating}
\vspace{-0.3cm}
\end{table}

While Hard EM achieves impressive performance when reembedding from scratch and when training on only 200 or 500 examples, we
wonder whether this performance is due to the alternating optimization, to the multiple E-step updates per M-step update, or to the lack of sampling.  
We accordingly experiment with optimizing our VQ-VAE and CatVAE variants in an alternating way, allowing multiple inference network updates per update of the generative parameters $\btheta$.
We show the results
on the AG News dataset in Table~\ref{tab:alternating}. We find that alternating does generally improve the performance of VQ-VAE and CatVAE as well, though Hard EM performs the best overall when reembedding from scratch. Furthermore, because Hard EM requires no sampling, it is a compelling alternative to CatVAE. For all three methods, we find that doing 3 inference network update steps during alternating optimization performs no better than doing a single one, which suggests that aggressively optimizing the inference network is not crucial in our setting.  

\subsection{Compression}

We briefly discuss in what sense discrete latent representations reduce storage requirements. Given a vocabulary of size 30,000, storing a $T$-length sentence requires $T \log_2 30000 \approx 14.9T$ bits. Our models require at most $ML \log_2 K$ bits to represent a sentence, which is generally smaller, and especially so when using a global representation. It is also worth noting that storing a $d$-dimensional floating point representation of a sentence (as continuous latent variable approaches might) costs $32d$ bits, which is typically much larger.

While the above holds for storage, the space required to \textit{classify} a sentence represented as $ML$ integers using a parametric classifier may not be smaller than that required for classifying a sentence represented as a $d$-dimensional floating point vector. On the other hand, nearest neighbor-based methods, which are experiencing renewed interest~\citep{guu2018generating,chen2019multi,wiseman2019label}, should be significantly less expensive in terms of time and memory when sentences are encoded as $ML$ integers rather than $d$-dimensional floating point vectors. In the next subsection we quantitatively evaluate our discrete representations in a nearest neighbor-based retrieval setting.

\subsection{Nearest Neighbor-Based Retrieval}

\begin{table}[t!]
\centering
\small

\input{tables/retrieval.tex}
\caption{
Unsupervised document retrieval on AG News dataset, measured by average label precision of top 100 nearest neighbors of the development set. \underline{Underlined} score is the row best.
Discrete representations use Hamming distance.
}
\label{tab:retrieval}
\vspace{-0.2cm}
\end{table}

In the classification experiments of Section~\ref{sec:evaluating}, we evaluated
our discrete representations by training a small classifier on top of them. Here we evaluate our global discrete representations in a document retrieval task to directly assess their quality; we note that this evaluation does not rely on the learned code books, embeddings, or a classifier. 

In these experiments we use each document in the development set of the AG News corpus as a query to retrieve $100$ nearest neighbors in the training corpus, as measured by Hamming distance. We use average label precision, the fraction of retrieved documents that have the same label as the query document, to evaluate the retrieved neighbors. We compare with baselines that use averaged $300d$ pretrained word vectors (corresponding to each token in the document) as a representation, where neighbors are retrieved based on cosine or $\mathcal{L}_2$ distance. We use GloVe with a 2.2 million vocabulary \citep{pennington-etal-2014-glove} and fastText with a 2 million vocabulary \citep{mikolov-etal-2018-advances}. The results are in Table~\ref{tab:retrieval}. We see that CatVAE and Hard EM outperform these CBOW baselines (while being significantly more space efficient), while VQ-VAE does not. These results are in line with those of Figure~\ref{fig:devresults}, where VQ-VAE struggles when its code book vectors cannot be used (i.e., when reembedding from scratch).

\begin{figure}[t!] 
\includegraphics[scale=0.14]{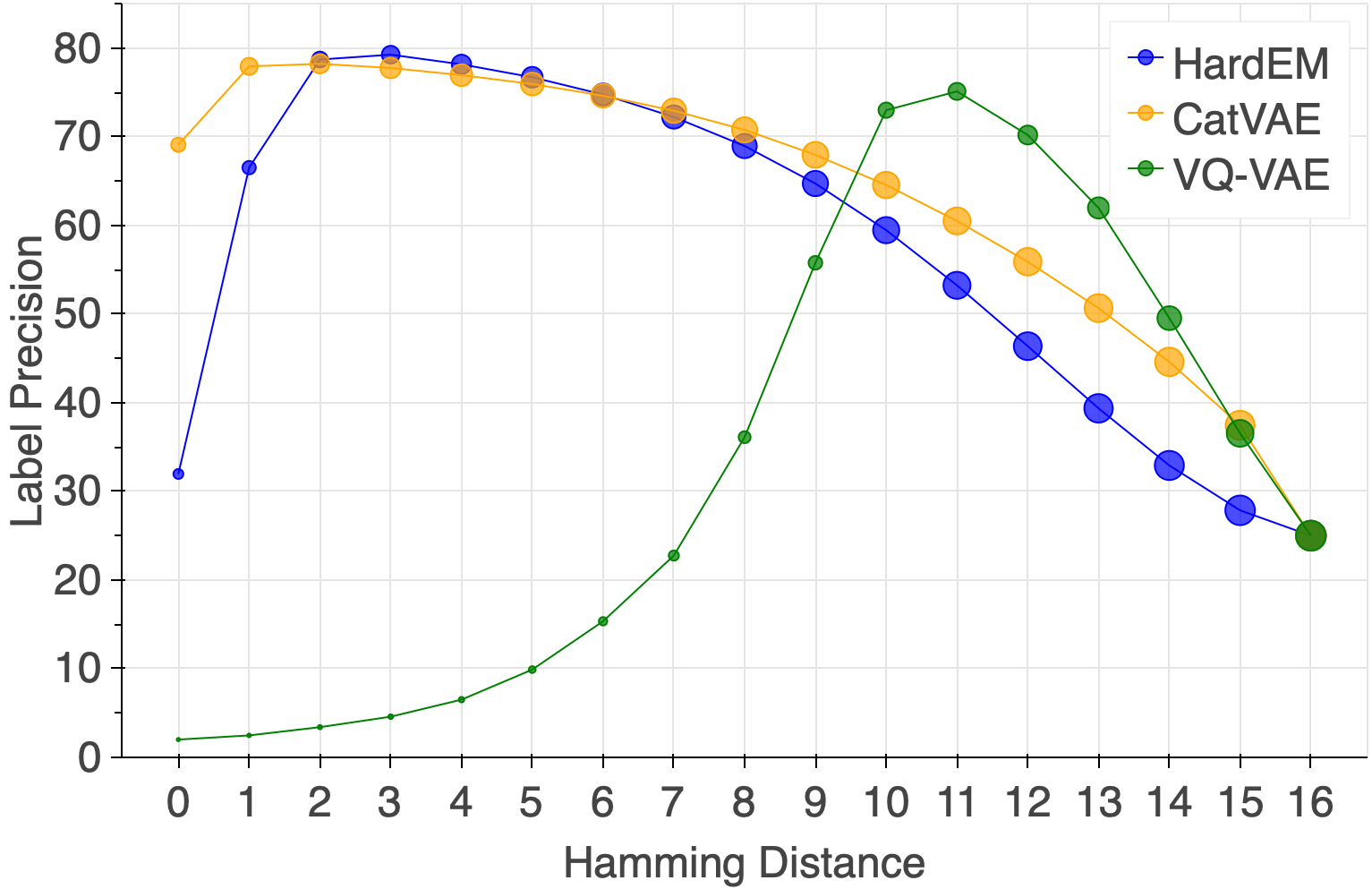}
\caption{Retrieving document clusters with Hamming distance $\le D$, for global models with $M=16$ and $K=256$. Query and target documents are from AG News's development set and training set respectively. Dot size indicates the number of documents in a cluster. 
}
\label{fig:m16}
\vspace{-0.2cm}
\end{figure}

In Figure~\ref{fig:m16} we additionally experiment with a slightly different setting: Rather than retrieving a fixed number of nearest neighbors for a query document, we retrieve all the documents within a neighborhood of Hamming distance $\le D$, and calculate the average label precision. These results use global representations with $M=16$, and we therefore examine thresholds of $D \in \{0, \ldots, 16\}$. We see that for CatVAE and Hard EM, the document similarity (or label precision) has an approximately linear correlation with Hamming distance. On the other hand, VQ-VAE shows a more surprising pattern, where high precision is not achieved until $D=10$, perhaps suggesting that a large portion of the latent dimensions are redundant.

\section{Conclusion}
We have presented experiments comparing the discrete representations learned by a Categorical VAE,
a VQ-VAE, and Hard EM in terms of their ability to improve a low-resource text classification system, and to allow for nearest neighbor-based document retrieval.
Our best classification models are able to outperform previous work, and this remains so even when we reembed discrete latents from scratch in the learned classifier.
We find that amortized Hard EM is particularly effective in low-resource regimes when reembedding from scratch, and that VQ-VAE struggles in these settings.

\section*{Acknowledgments}
This material is based upon work supported by the Air Force Office of Scientific Research under award number FA9550-18-1-0166.

\bibliography{acl2020}
\bibliographystyle{acl_natbib}

\input{appendix}

\end{document}

%% file: mathdefs.tex
\newcommand{\boldh}{\mathbf{h}}

\newcommand{\bolde}{\mathbf{e}}

\newcommand{\boldW}{\mathbf{W}}

\newcommand{\btheta}{\boldsymbol{\theta}}

\newcommand{\bphi}{\boldsymbol{\phi}}

\newcommand{\E}{\mathbb{E}}

\newcommand{\reals}{\mathbb{R}}

\DeclareMathOperator*{\KL}{KL}
\DeclareMathOperator*{\ELBO}{ELBO}

\DeclareMathOperator*{\softmax}{softmax}

\DeclareMathOperator*{\enc}{enc}

\newcommand{\given}{\,|\,}
\newcommand{\param}{;}

%% file: tables/tasks.tex
\begin{tabular}{lrrrr}
\toprule
\textbf{Dataset} & \textbf{\# Classes} & \textbf{Train} & \textbf{Dev} &  \textbf{Test}\\
\midrule
AG News & 4 & 115K & 5K & 7.6K\\
DBPedia & 14 & 555K & 5K & 70K\\
Yelp Review Full & 5 & 645K & 5K & 50K\\
\bottomrule
\end{tabular}

%% file: tables/hypers.tex
\begin{tabular}{llrr}
\toprule
& Method & $K$ & $M$ \\
\midrule
Local & CatVAE & 4096 & 1 \\
Local (re) & Hard EM & 1024 & 1 \\ %
Global  & CatVAE & 256 & 4 \\
Global (re)  & Hard EM & 4096 & 4 \\ %
\bottomrule
\end{tabular}

%% file: tables/test.tex
\begin{tabular}{@{}l@{\hspace{1ex}}cccc@{}}
\toprule
Model & 200 & 500 & 2500 & Full\\
\midrule
\multicolumn{5}{c}{AG News} \\
\midrule
CBOW & 63.4 (1.5) & 72.9 (0.7) & 82.1 (0.2) & 90.0 (0.2) \\
VAMPIRE$^\star$ & 83.9 (0.6) & 84.5 (0.4) & 85.8 (0.2) & - \\
VAMPIRE & 82.2 (0.8) & 84.7 (0.2) & \textbf{86.4} (0.4) & \textbf{91.0} (0.1) \\ 
Local & 82.7 (0.1) & 84.3 (0.3) & 85.0 (0.4) & 86.6 (0.2) \\
Local (re) & 82.7 (0.4) & 84.0 (0.3) & 85.4 (0.1) & 87.1 (0.3) \\
Global & \textbf{84.6} (0.1) & \textbf{85.7} (0.1) & 86.3 (0.2) & 87.5 (0.6) \\
Global (re) &83.9 (0.5) & 84.6 (0.2) & 85.1 (0.3) & 86.9 (0.1) \\
\midrule
\multicolumn{5}{c}{DBPedia} \\
\midrule
CBOW &72.7 (0.6) & 84.7 (0.7) & 92.8 (0.3) & 97.7 (0.1) \\
VAMPIRE & 89.1 (1.3) & 93.7 (0.5) & \textbf{95.7} (0.2) & \textbf{98.2} (0.1) \\ 
Local &89.2 (0.2) & 92.8 (0.4) & 94.6 (0.2) & 97.1 (0.3) \\
Local (re) &88.7 (0.2) & 90.2 (0.3) & 93.3 (0.1) & 96.9 (0.2) \\
Global &\textbf{91.8} (0.5) & \textbf{94.3} (0.3) & 95.0 (0.2) & 95.6 (0.0) \\
Global (re) & 88.5 (0.7) & 92.3 (0.7) & 94.6 (0.4) & 95.8 (0.1) \\
\midrule
\multicolumn{5}{c}{Yelp Full} \\
\midrule
CBOW & 31.0 (5.9) & 41.1 (0.6) & 48.4 (0.4) & 58.9 (0.4) \\ 
VAMPIRE & 41.4 (2.9) & 47.2 (0.7) & 52.5 (0.1) & \textbf{60.3} (0.1) \\ 
Local & 46.2 (0.8) & 49.0 (0.5) & 51.9 (0.5) & 53.1 (0.3) \\
Local (re) &47.2 (0.7) & 49.4 (0.7) & 52.1 (0.2) & 55.0 (0.6) \\
Global &\textbf{48.5} (1.0) & \textbf{50.1} (0.5) & \textbf{53.0} (0.3) & 54.9 (0.4) \\
Global (re) & 46.0 (0.5) & 47.4 (0.5) & 48.8 (0.8) & 53.8 (0.3) \\
\bottomrule
\end{tabular}

%% file: tables/examples.tex
\begin{tabular}{@{}lll@{}}
\toprule
\textbf{Cluster} & \textbf{Class} & \textbf{Text}\\
\midrule
(23, 155, 24, 53) & World & a platoon in iraq is being investigated for allegedly refusing to carry out a convoy mission... \\ %
& World & afp chechen warlord shamil basayev has claimed responsibility for the deadly school... \\ %
& World & the federal government has sent a team of defence personnel to verify a claim that two...\\
& World & an audio tape purportedly by osama bin laden praises gunmen who attacked a us consulate...\\
\midrule
(41, 75, 175, 222) & Business & amazon com says it has reached an agreement to buy joyo com, the largest internet retailer...\\ %
& Business & electronic data systems offered voluntary early retirement to about 9, 200 us employees...\\
& Business & in the aftermath of its purchase of at amp t wireless, cingular wireless is selling several sets...\\ %
& Sci/Tech &  wired amp wireless continues its reign at the top spot among it priorities due to widespread...\\
\midrule
(10, 208, 179, 180)  & Sports & this is the week of the season when every giants defensive back needs to have shoulders as...\\
&Sports & drew henson will have to wait before he's the star of the dallas cowboys offense right now...\\
&Sports & st louis how do you beat the greatest show on turf with two rookie cornerbacks...\\ %
& Sports & cincinnati bengals coach marvin lewis said yesterday that he expects quarterback carson...\\
\midrule
(65, 224, 78, 114) & Sci/Tech & microsoft acknowledged on monday it continued to battle a technical glitch that prevented...\\
& Sci/Tech & users of the music player should watch out for hacked themes a flaw allows would be...\\
& World & microsoft's popular internet explorer has a serious rival in the firefox browser\\
& Sci/Tech & microsoft has doubled the period of time it will allow business users of windows xp to...\\
\bottomrule
\end{tabular}

%% file: tables/examples2.tex
\begin{tabular}{@{}ll@{}}
\toprule
\textbf{Cluster} & \textbf{Words}\\
\midrule
822 & government indonesia guilty prison general prosecutors leader law german sex authorities charged marched issue \\
651 & yankees veteran baltimore quarterback offense tampa steelers giants defensive cleveland minnesota pittsburgh \\ %
595 & month currency low session dollar euro greenback yen monetary weakening lows versus maintained grip rebounded \\ %
305 & if despite when although \\
304 & core plans intel athlon opteron processors chip hewlett packard strategy clearer forum designs desktop upped ante \\ %
802 & bit cameras image pleasing integrates multimedia functions gprs automation self types btx supercomputers logic \\ %
298 & president dick cheney john republicans kerry voters democrat javier sen kellogg \\
994 & exploded bomb near killing injuring explosion eight residents firefighters leak central philippine 55 heavily cancun \\ %
484 & apple atari san francisco sony toshiba anaheim finally assault famed mp3 freedom u2 accusations brook introduces\\ %
510 & iraq killed car rebel iraqi military suicide forces marines insurgents baghdad evacuation bomber strikes explosions \\ %
\bottomrule
\end{tabular}

%% file: tables/alternating.tex
\begin{tabular}{@{}l@{\hspace{2ex}}cccc@{}}
\toprule
\textbf{Model} & \textbf{200} & \textbf{200 (re)} & \textbf{500} & \textbf{500 (re)} \\
\midrule
EM-Local & 81.4 & 82.1 & 83.0 & 82.8 \\ 
EM-Global & 85.6 & \textbf{84.6} & 85.5 & \textbf{85.4}\\
\midrule
Cat-Local-Alt & 83.3 & 82.9 & 84.8 & 84.1 \\ 
Cat-Global-Alt & \textbf{86.4} & 83.1 & \textbf{87.1} & 85.0\\
\midrule
Cat-Local & 83.2 & 82.5 & 85.3 & 84.8 \\ 
Cat-Global & 85.4 & 82.8 & 86.1 & 84.5\\
\midrule
VQ-Local-Alt & 82.9 & 81.1 & 84.8 & 81.4 \\ 
VQ-Global-Alt & 84.7 & 79.6 & 85.9 & 82.9\\
\midrule
VQ-Local & 82.6 & 78.7 & 83.6 & 81.3 \\ 
VQ-Global & 83.0 & 76.8 & 85.4 & 82.0\\
\bottomrule
\end{tabular}

%% file: tables/retrieval.tex
\begin{tabular}{@{}l@{}ccc@{}}
\toprule
& \multicolumn{3}{c} {Discrete Embedding} \\
& {M=4, K=256} & {M=8, K=128} & {M=16, K=256} \\
\midrule
Hard EM & {76.1} & {\underline{79.6}} & {78.8} \\ 
CatVAE & {77.5} & {73.7} & {\underline{78.5}} \\ 
VQ-VAE & {69.1} & {\underline{73.5}} & {71.2} \\ 
\midrule
& \multicolumn{3}{c}{Continuous Embedding ($300d$)} \\
& {$\mathcal{L}_2$} & {\textsc{Cosine}} &   \\
\midrule
GloVe & {76.4} & {\underline{76.6}} & \\
fastText & {72.8} & {\underline{74.1}} & \\
\bottomrule
\end{tabular}

%% file: appendix.tex
\appendix

\section{Model Comparison by Datasets}\label{sec:supplemental}

\begin{figure*}[t!]
\centering
\includegraphics[scale=0.65]{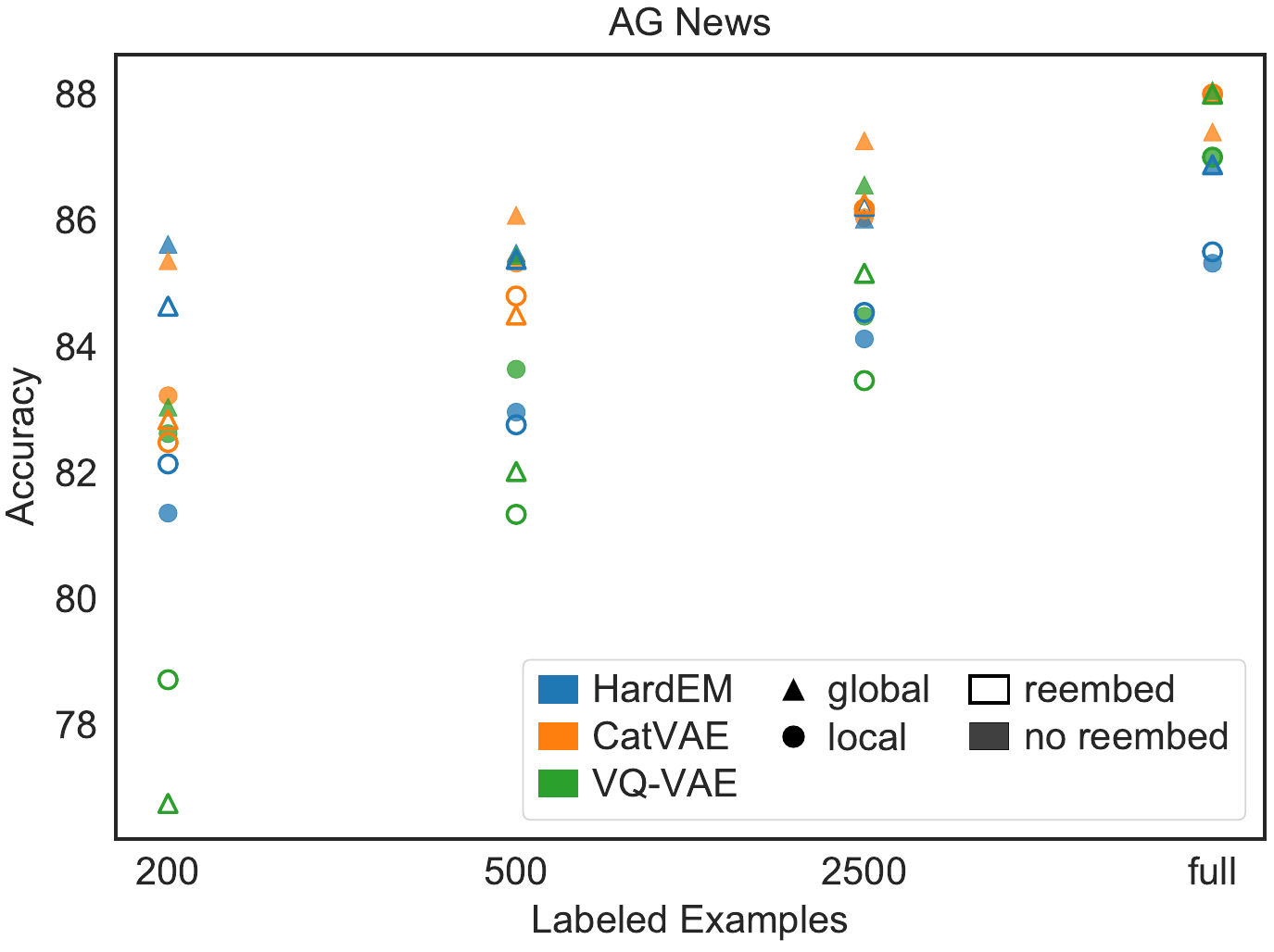}
    
\medskip
    
\includegraphics[scale=0.65]{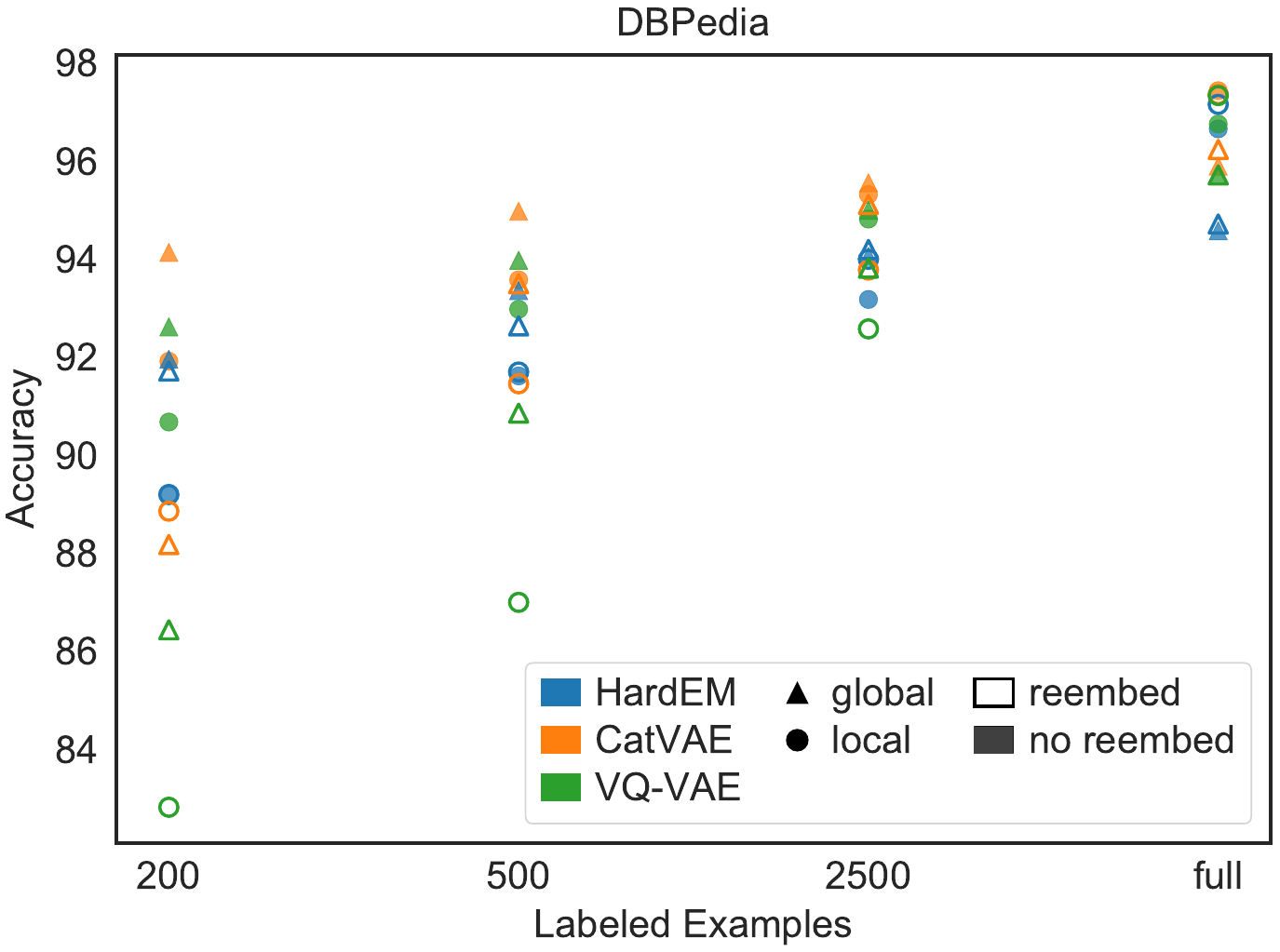}
    
\medskip
    
\includegraphics[scale=0.65]{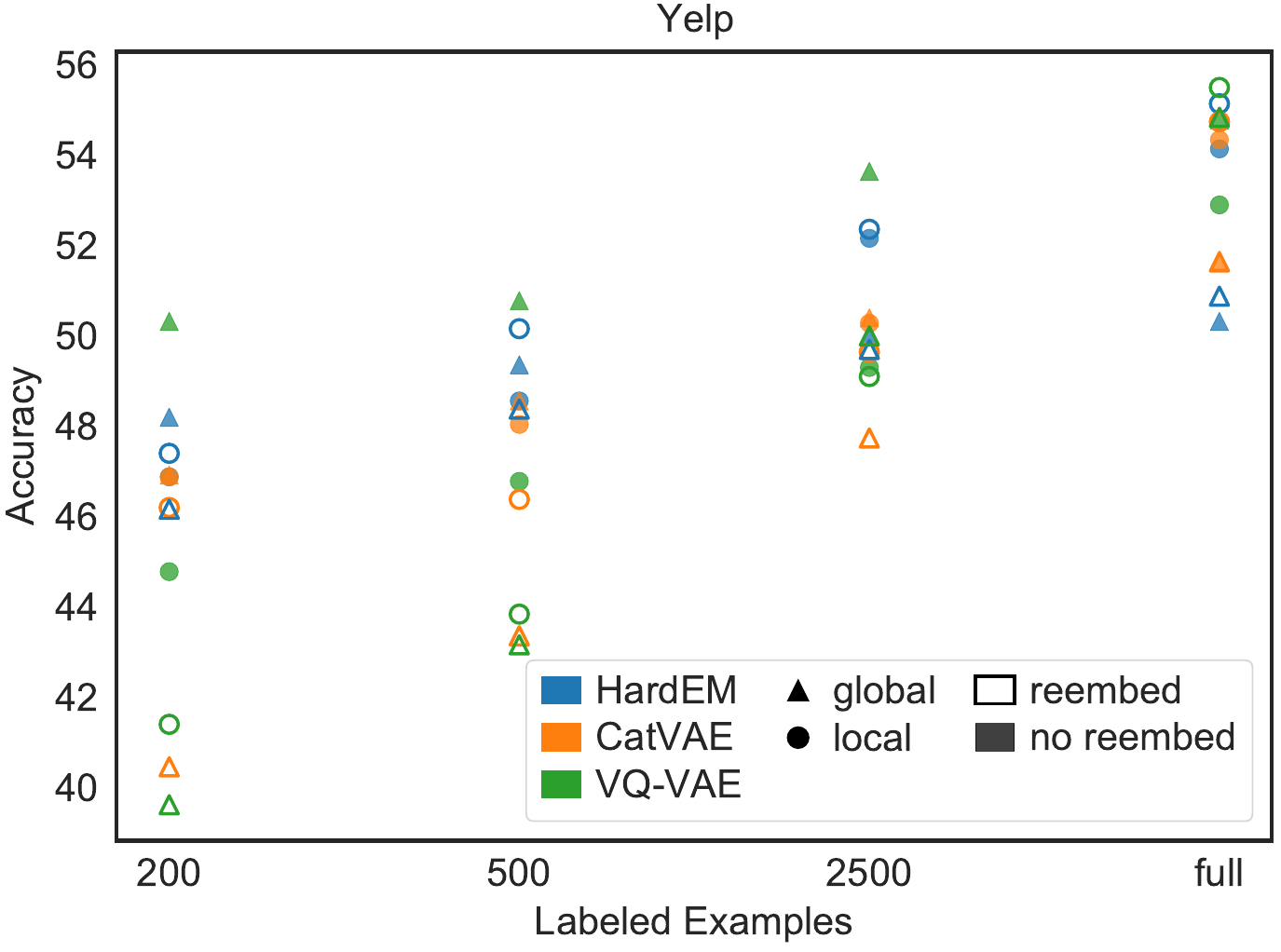}
    
\caption{The accuracies obtained by Hard EM, Categorical VAE, and VQ-VAE representations on
the development datasets of AG News (top), DBPedia (middle), and Yelp Full (bottom),
for different numbers of labeled training examples. Triangular and circular markers correspond to global and local models, respectively. Unshaded and shaded markers correspond to reembedding from scratch and using encoder embeddings, respectively.
}
\label{fig:suppresults}
\end{figure*}

We plot the development set classification performance of each method, this time distinguishing between datasets, in Figure~\ref{fig:suppresults}.

